\documentclass[runningheads]{llncs}

\usepackage{eccv}

\usepackage{eccvabbrv}

\usepackage{graphicx}
\usepackage{booktabs}
\usepackage{todonotes}
\usepackage[accsupp]{axessibility}  % Improves PDF readability for those with disabilities.

\usepackage[pagebackref,breaklinks,colorlinks,citecolor=eccvblue]{hyperref}
\usepackage{orcidlink}
\usepackage{rotating}
\usepackage{makecell}

\begin{document}

\title{Fast Occupancy Network} 


\author{Mingjie Lu \and
Yuanxian Huang \and
Ji Liu \and
Xingliang Huang \and
Dong Li \and
Jinzhang Peng \and
Lu Tian \and
Emad Barsoum}

\authorrunning{Lu \textit{et al}.}

\institute{Advanced Micro Devices, Inc., Beijing, China\\
\email{\{mingjiel, yxhuang, liuji, d.li, jinz.peng, lu.tian, ebarsoum\}@amd.com}}

\maketitle

\begin{abstract}
Occupancy Network has recently attracted much attention in autonomous driving. Instead of monocular 3D detection and recent bird's eye view(BEV) models predicting 3D bounding box of obstacles, Occupancy Network predicts the category of voxel in specified 3D space around the ego vehicle via transforming 3D detection task into 3D voxel segmentation task, which has much superiority in tackling category outlier obstacles and providing fine-grained 3D representation. However, existing methods usually require huge computation resources than previous methods, which hinder the Occupancy Network solution applying in intelligent driving systems. To address this problem, we make an analysis of the bottleneck of Occupancy Network inference cost, and present a simple and fast Occupancy Network model, which adopts a deformable 2D convolutional layer to lift BEV feature to 3D voxel feature and presents an efficient voxel feature pyramid network (FPN) module to improve performance with few computational cost. Further, we present a cost-free 2D segmentation branch in perspective view after feature extractors for Occupancy Network during inference phase to improve accuracy.  Experimental results demonstrate that
our method consistently outperforms existing methods in both accuracy and inference speed, which surpasses recent state-of-the-art (SOTA) OCCNet by 1.7\% with ResNet50 backbone with about 3$\times$ inference speedup. What's more, our method can be easily applied to existing BEV models to transform them into Occupancy Network models.
\end{abstract}  

\section{Introduction}

Perception system plays an important role in modern intelligent or autonomous driving, which usually detects various obstacles and road layout information and transmits these surrounding environmental information to downstream planning module. In the early stage, 2D detection models~\cite{redmon2018yolov3,liu2016ssd}, segmentation models~\cite{YuanCW19,chen2017rethinking}, and lane line detection models~\cite{tabelini2021keep,qu2021focus} are leveraged to build the basic perception systems. With LiDAR-based 3D detection methods~\cite{lang2019pointpillars,yin2021center} being introduced to intelligent driving, the perception ability has a significant increase, which leads to a giant leap to autonomous driving. Further the sensor fusion methods~\cite{vora2020pointpainting,xu2021fusionpainting} combine several observations from multiple sensors to obtain more robust and accurate perception results. However, due to high cost of LiDAR sensors, vision-based perception methods attract much attention and almost adopt BEV representation as a mainstream way in the intelligent driving field over the past years.

Vision-based BEV methods usually take multi-view camera images as inputs and leverage various strategies to perform 2D-to-3D transformation ~\cite{bevdet,li2022bevformer} and forward the BEV features to various task heads, which usually include 3D detection~\cite{huang2021bevdet,bevdepth,li2022bevformer,liu2023petrv2}, BEV segmentation\cite{peng2023bevsegformer}, and high-definition (HD) map construction~\cite{maptr,maptrv2,li2022hdmapnet,liu2023vectormapnet}. The 3D object detection task focuses on representing objects with rigid 3D bounding boxes, which is difficult to depict the various shapes of obstacles in the real world.  
BEV semantic segmentation only requires to predict the semantic map at the horizontal level, which provides limited semantic information for intelligent driving.
The HD map construction task primarily emphasizes the static elements on the road(i.e., lane divider, road boundaries, etc).
However, these tasks only represent the 3D environment in a limited perspective without holistic understanding of the 3D scene. The 3D  semantic occupancy~\cite{tpvformer,occformer,sima2023_occnet} has been proposed to represent the surrounding environmental information at a finer granularity, which can depict obstacles with various shapes, road layout information, and other elements with predefined 3D voxels in real driving scenarios. 

The 3D semantic occupancy is also known as 3D semantic scene completion (SSC)~\cite{song2017semantic} with taking LiDAR point cloud as input. LiDAR-based methods ~\cite{roldao2020lmscnet,cheng2021s3cnet,chen20203d,li2020anisotropic} have achieved outstanding performance resulting from explicit depth measurements of point cloud. Owing to the expensive cost of LiDAR sensors, vision-based 3D occupancy has attracted much attention, which takes multi-view camera images as input with various methods to transform multi-view inputs into voxel-based feature representation for 3D occupancy prediction. TPVFormer~\cite{tpvformer} obtains the voxel features by representing the 3D space by generalizing the BEV to tri-perspective views. Specifically, a 3D point can be projected into three panels and a voxel feature can be acquired by bilinearly interpolating on the tri-perspective features. Such a decoupled representation allows low lifting cost but may show performance decrease when complete environment features are required.
Meanwhile, some methods\cite{sima2023_occnet,panoocc,occformer} construct the voxel features by directly generalizing the voxel queries. However, due to the huge search space, all the 3D voxel representation incurs high memory costs and significantly impedes inference speed.
Therefore, the solution to effective 3D voxel representation lies in acquiring efficient and effective 3D information while minimizing computational cost.  OccNet~\cite{sima2023_occnet} presents a simple and fast BEVNet, which lifts BEV features to 3D voxel features with a simple MLP layer and obtains a not bad performance with 2 points gap to SOTA method~\cite{sima2023_occnet}. This phenomenon shows the BEV feature contains quite 3D information in vertical direction and has the potential of achieving higher accuracy with more exploration. %To this end, we rethink how far the single-panel BEV features can reach in the voxel representing for the occupancy semantic prediction task.
We verify several methods involving 2D convolution and 2D deformable convolution via expanding receptive field in the lifting process and 3D deformable cross-attention method to find the best method. Finally, we lift the BEV features to 3D voxel features by a deformable convolutional layer without usual 3D cross-attention modules for BEV feature lifting, which requires less computational cost than existing methods. We prove that this lifting method is able to recover voxel information from the BEV feature and achieve comparable performance with complex lifting modules from the existing method. Based on this finding, a fast and strong 3D occupancy network is proposed, and it achieves about 3$\times$ inference speedup compared to existing SOTA method OccNet~\cite{sima2023_occnet} on ResNet50 backbone with our proposed multi-perspective-view segmentation supervision and Partial Voxel FPN.
Our main contributions can be summarized as follows.
(1) We propose a simple yet efficient occupancy network based on BEV features and lift with a deformable convolution layer. (2) We propose an efficient voxel FPN module for 3D voxel features to improve performance. (3) We present a cost-free 2D segmentation branch in perspective view after feature extractors for occupancy network during the inference phase to improve accuracy  (4) Comprehensive experiments are conducted on both accuracy and inference speed, demonstrating the superiority of our method.

%5. Issues of existing methods and our solution

%6. Contribution summary

\section{Related Work}
% \subsection{Bev-based Methods}
\subsection{BEV Methods}

The BEV has been extensively used for trajectory prediction and control. The general representation capability of BEVs for 3D space has recently led to considerable progress in many 3D perception tasks(e.g., 3D object detection~\cite{li2022bevformer, bevdepth}, BEV semantic segmentation~\cite{bevseg}, and online high-definition map construction~\cite{maptr, pivotnet}).
BEV features have been crucial in advancing camera-based 3D perception methods. The view transformation is a key process in BEV feature construction. Many methods utilize the depth estimation information to transform the perspective features to BEV space. For example, the Lift-Splat-Shoot~\cite{philion2020lift} method splats image features from multi-view cameras onto a BEV grid, determined by the depth distribution of each pixel in the image. The acquired BEV features are capable of executing segmentation or detection tasks in a 3D environment. 
BEV-Seg~\cite{bevseg} performs semantic segmentation and depth estimation for each view before view transformation, and then uses a parse network for BEV semantic segmentation prediction.
VPN~\cite{vpn} utilizes a multi-layer perceptron (MLP) for the direct prediction of BEV segmentation from multi-view images.
Another kind of view transform method adopts an attention mechanism to directly acquire features on the multi-view images using a set of BEV queries. BEVFormer~\cite{li2022bevformer} designs a spatial-cross attention module to encode the multi-camera features into BEV space via pre-defined BEV features. Similarly, MapTR~\cite{maptr} also uses deformable attention in a Transformer encoder to acquire BEV features, which are decoded to a set of map elements by a DETR-like~\cite{detr} decoder.

\subsection{Occupancy Network}
Occupancy prediction expands the BEV space in the height dimension to explicitly representing each voxel in the 3D space. Occupancy prediction can provide accurate 3D information for planning in autonomous driving, as it accurately depicts an object's geometrical structure. Generally, existing methods achieve occupancy prediction in the BEV space or voxel space. 

\noindent{\bf{BEV-based Methods.}}
Migrating the BEV perception task to occupancy doesn't appear to be a challenging task. It merely requires determining how to lift the BEV feature to 3D and identify the occupation head.
MonoScene~\cite{monoscene} first employs U-Net to infer dense 3D occupancy with semantic labels from a single monocular RGB image. 
BEVDet~\cite{bevdet} can directly predict occupancy by replacing the detection head with the scene completion head, which is constructed on the BEV feature maps.
TPVFormer~\cite{tpvformer} suggests a tri-perspective view approach for estimating 3D occupancy. They merge the tri-perspective view feature, where each voxel is situated, to accomplish the 2D-to-3D transformation.

\noindent{\bf{Voxel-based Methods.}}
FB-OCC~\cite{li2023fbocc} is a novel method that combines voxel-BEV representation using forward-backward feature projection. By projecting features in both forward and backward directions, FB-OCC can better understand the context and interactions. 
VoxFormer~\cite{voxformer} employs depth estimation to establish voxel queries and then a deformable cross-attention captures the 3D structure of the scene.
OccDepth~\cite{occdepth} also adopts a depth-aware method to predict semantic occupancy and a distillation is utilized to improve performance.
OccFormer~\cite{occformer} uses a dual-path transformer network, the combination of voxel and bev features can provide local and global information along the horizontal plane.
OccNet~\cite{sima2023_occnet} also uses both BEV cross-attention and voxel cross-attention to achieve a trade-off in performance and efficiency.
PanoOcc~\cite{panoocc} uses 3D voxel queries with a coarse-to-fine learning scheme to reduce the complexity of 3D cross-attention. 

\subsection{Feature Pyramid Network}

The feature pyramid network (FPN)~\cite{fpn} is proposed to address the multi-scale problem in 2D object detection.
FPN fuses the rich semantic information of the higher-level features in the feature pyramid with the location information of the lower-level features. It solves the multi-scale problem in object detection by making predictions on feature maps of different scales separately.
Inspired by 2D FPN, Voxel-FPN~\cite{voxelfpn} utilizes multi-layer 3D convolution to extract voxel features into multi-scale features in a bottom-up fashion, and then performs 3D object detection at multiple scales. 
The multi-scale fusion strategy of FPN can enhance the Occupancy network's voxel segmentation performance as well.
In FB-OCC~\cite{li2023fbocc}, the occupancy prediction header employs a 3D FPN for decoding the voxel features. By employing 3D convolution, it extracts high-dimensional semantic voxel features to enhance the low-dimensional voxel features iteratively.

Our method belongs to the BEV-based methods, and unlike the existing methods using 3D deformable attention, we directly lift the BEV features to voxel features by deformable convolution in BEV space. This significantly reduces the query number and the computational cost of constructing voxel features. To improve the semantic segmentation performance of Occupancy, we design a Partial Voxel FPN structure. Unlike Voxel-FPN~\cite{voxelfpn}, which fuses multi-scale voxel features at all levels, our approach merges multi-scale features exclusively in partial height layers. This achieves similar multi-scale fusion goals as FPN while reducing computational expenses.

\begin{figure*}[t]
    % \vspace{-8mm}
    \centering
    \begin{minipage}{1\linewidth}
        \centering
        \begin{subfigure}{1\linewidth}
        \includegraphics[width=1\textwidth]{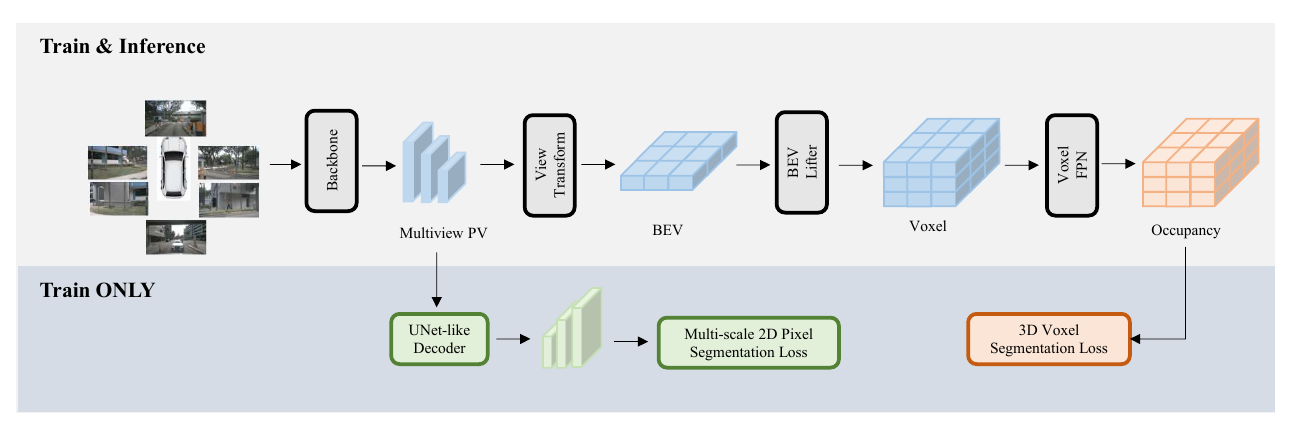}
        \end{subfigure}
    \end{minipage} 
    \vspace{-2mm}
    \caption{Overview of our Fast Occupancy Network pipeline.}
    \vspace{-8mm}
    \label{fig:overall_framework}
    
\end{figure*}

\section{Method}
\subsection{Overview}
% 1. method description  \\
%2. overview figure
The overall pipeline of our Fast Occupancy Network is shown in Fig.~\ref{fig:overall_framework}.
First, the image encoder extracts the multi-view features from the raw camera images. The multi-view features consist of a multi-scale feature pyramid that provides semantic features at different resolutions.
Then, the image-to-BEV method transforms the multi-view features into the BEV space using deformable cross-attention.
Next, the BEV features are lifted to 3D voxel space using a simple but efficient deformable convolutional layer in our occupancy decoder.
The occupancy prediction is performed in the 3D voxel space, and the 3D voxel features are further decoded by the proposed Partial Voxel FPN.
Last, the voxel predictions are supervised by voxel-wise classification loss during training.
The voxel predictions can be transformed into other 3D environmental perception tasks, such as lidar segmentation.
Our occupancy network is quite simple but efficient based on BEV features and performs lifting to 3D voxel features with a convolutional layer. Specifically, we first introduce the image encoder and image-to-bev transformation in this section. Next, we will illustrate our proposed module in Sec~\ref{sec_modul} including the BEV lifting module, Partial Voxel FPN, and perspective-view (PV) supervision.

\noindent{\bf{Image Encoder.}}
Image Encoder is used to extract 2D semantic features from multi-view camera images.
We utilize ResNet50 and ResNet101 as backbones to extract deep features in 2D space.
Taking any camera images $I \in \mathbb{R}^{N \times 3 \times H \times W}$ as input, the image encoder generates feature pyramids $P_i \in \mathbb{R}^{N \times C_i, \times H_i \times W_i}$, where $N$ denote the number of perspective views, $H$ and $W$ denotes the input size of camera images, $C$ is the channel number, and the subscript $i$ denotes the corresponding variable in the $i$-th layer.

\noindent{\bf{Image-to-BEV Transformation.}}
Following~\cite{li2022bevformer}, we use BEV queries to transform 3D reference points into 2D perspective views. Features that hit in the 2D perspective space are weighted to obtain the corresponding features in the BEV space through a spatial cross-attention mechanism.
The obtained BEV features can be used by subsequent occupancy decoders to obtain 3D voxel feature which is used to further predict semantic categories in 3D space.

\begin{figure}[t]
    \centering
    \begin{minipage}{0.7\linewidth}
        \centering
        \begin{subfigure}{1\linewidth}
        \includegraphics[width=1\textwidth]{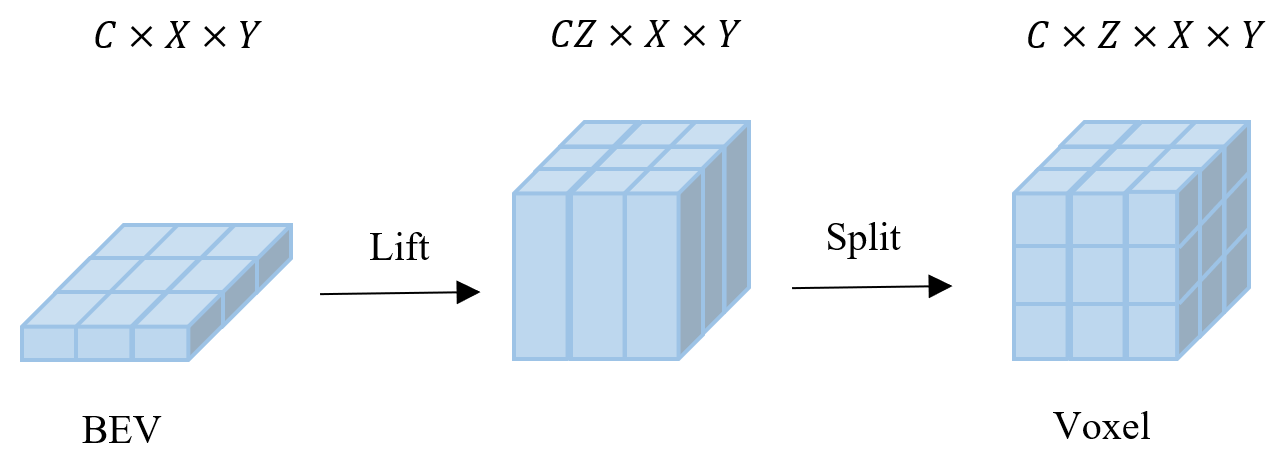}
        \end{subfigure}
    \end{minipage} 
     \vspace{-4mm}
    \caption{The structure of the BEV lifter module. }
    \vspace{-6mm}
    \label{fig:module-bevlifter}
\end{figure}

\subsection{Module Details}
\label{sec_modul}
\noindent{\bf{BEV lifting methods.}} The 2D deformable cross-attention method ~\cite{zhu2020deformable} solves the problem of view-transformation from the image feature to the BEV feature very well. Given an input feature map $x \in \mathbb{R}^{C \times H_{f} \times W_{f}}$ , a 2D query $q$ and a 2D reference point $p$, the deformable attention feature is calculated by
\begin{equation}
DeformAttn(q, p, x) =\sum_{m=1}^{M}W_{m} \sum_{k=1}^{K}A_{mqk}W_{m}' x (p+\Delta p_{mk}),
\end{equation}
where $m$ indexes the attention head, $k$ indexes the sampled keys, and $K$ is the total sampled key
number ($K \ll H_{f}W_{f}$). $W_{m}$ and $W_{k}$ are the learning weights, $A_{mqk}$ is the attention weight, and $p+\Delta p_{mk}$ is the learnable sample point position.
% 补充A和W的定义
The occupancy task requires obtaining the 3D features and classifying each voxel into predefined categories. Some methods~\cite{bevdet, huang2022bevdet4d} lift the bev features obtained by the 2D deformable attention module to 3D features by using MLP to expand the channel and then reshape to the height dimension. 
Others ~\cite{voxformer,panoocc} directly obtain 3D features by initializing 3D queries and positional embeddings, then extract 3D features with 3D deformable attention.

We find that the difference between 2D deformable cross attention and 3D deformable cross attention is the number of queries and whether or not they collapse height at the end, while the reference points and reference features are the same. 
However, the 3D deformable attention is too time-consuming due to the huge count of queries.
Meanwhile, the BEVNet experiments in OccNet \cite{sima2023_occnet} show the BEV features contain quite 3D information and have the potential to achieve higher performance. 
% In particular, we also find that when the number of 2D heads becomes $height$ times and the reduced sum op is removed, it becomes 3D. And so much query is unnecessary, there is a lot of room for optimization. 
It makes sense to replace 3D deformable attention by lifting bev features with efficient methods.
% We observe that the 3D output has much more to do with Value than Query, which is only responsible for generating attention weight and point offset. When we parameterize attention weight $A_{mk}$ and fix the reference point to the current pixel, the 3D deformable attention degenerates into a deformable convolution.

As shown in Figure~\ref{fig:module-bevlifter}, we simply obtain the voxel feature from the BEV feature in two steps. First, we `lift' the BEV feature up by increasing the dimensions of the channels. Then we `split' the increased channel dimension by the reshape operation to obtain the voxel features.
% Since the common method of using MLP to expand feature dimensions seems naive. It does not consider the relationship between surrounding pixels.
A common method to lift the BEV feature is using MLP. However, it does not consider the relationship between surrounding pixels. To overcome this limitation, we adopt 2D deformable convolution as our lifting method. On the one hand, it can expand the receptive field, and on the other hand, since the geometries of objects at different heights are inconsistent, dynamic offset is necessary.
In Sec~\ref{sec_ablation_uplift}, our experiments demonstrate that deformable convolution can achieve higher performance with high efficiency.

% voxel fpn figure
\begin{figure}[t]
    \centering
    \begin{minipage}{0.7\linewidth}
        \centering
        \begin{subfigure}{1\linewidth}
        \includegraphics[width=1\textwidth]{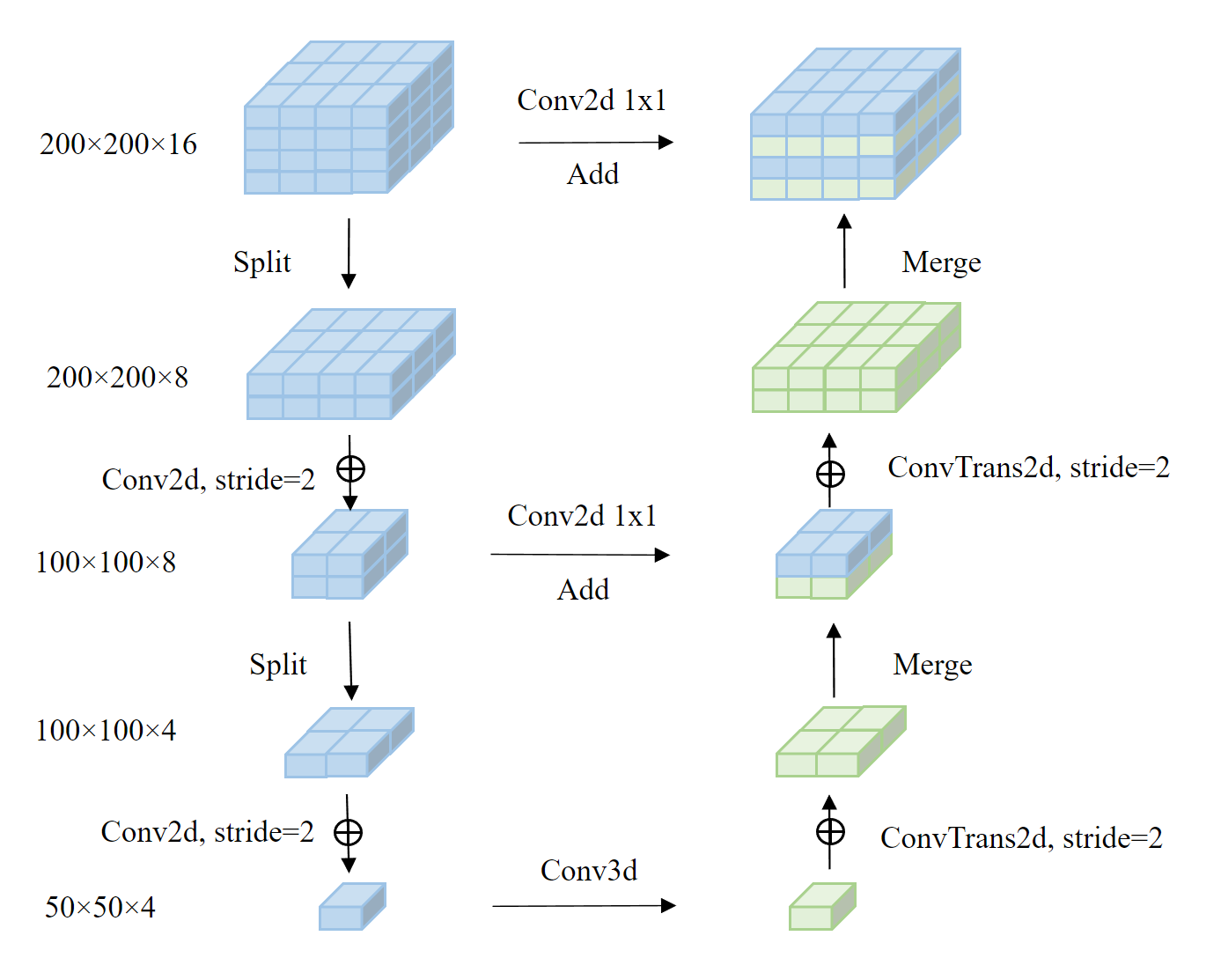}
        \end{subfigure}
    \end{minipage} 
     \vspace{-4mm}
    \caption{Overview of our Partial Voxel FPN.}
    \vspace{-6mm}
    \label{fig:module-voxelfpn}
\end{figure}

%2. Partial Voxel FPN \\
\noindent{\bf{Partial Voxel FPN.}}
Inspired by FB-OCC~\cite{li2023fbocc}, the multi-scale voxel feature fusion is key and important for 3D occupancy accuracy. However, using 3D convolution or deconvolution to realize the embedding of upscale and downscale is time-consuming. We propose a Partial Voxel FPN to realize the fusion of multi-scale features efficiently. General voxel FPN downsample xyz three-dimensions equally via 3D convolution. However, we strip out the heights and only perform a 2D convolution on the planar features of each height layer. 
In addition, in order to speed up voxel embedding, we sample half of the original feature at equal intervals, which will be kept intact, and only downsample and upsample the other half. The original feature will then be added to the upsampled feature. Finally, we use a 3D convolution on the minimum scale resolution to compensate for the lack of interactions on heights at which the cost is negligible.

Concretely, Figure~\ref{fig:module-voxelfpn} shows our Partial Voxel FPN framework. Firstly we split the original feature at the height, then downsample the xy-dimension by convolution with stride=2, which results in a feature that is downsampled in all three dimensions. Repeat this operation until you get the minimum scale feature. Secondly, a cheap 3D convolution acts on a feature size of $50\times50\times4$ on our implementation. Thirdly, a multi-scale feature contains two parts.
A portion is from the original feature, which is further passed with a $1\times1$ convolution. The other half is upsampled from the lower low-resolution feature and added to the other part of the original feature.
Sec.~\ref{sec_ablation_pvf} shows our method can dramatically improve inference speed while maintaining performance.

% Figure PV seg framework 
\begin{figure}[t]
    \centering
    \begin{minipage}{0.6\linewidth}
        \centering
        \begin{subfigure}{0.8\linewidth}
        \includegraphics[width=1\textwidth]{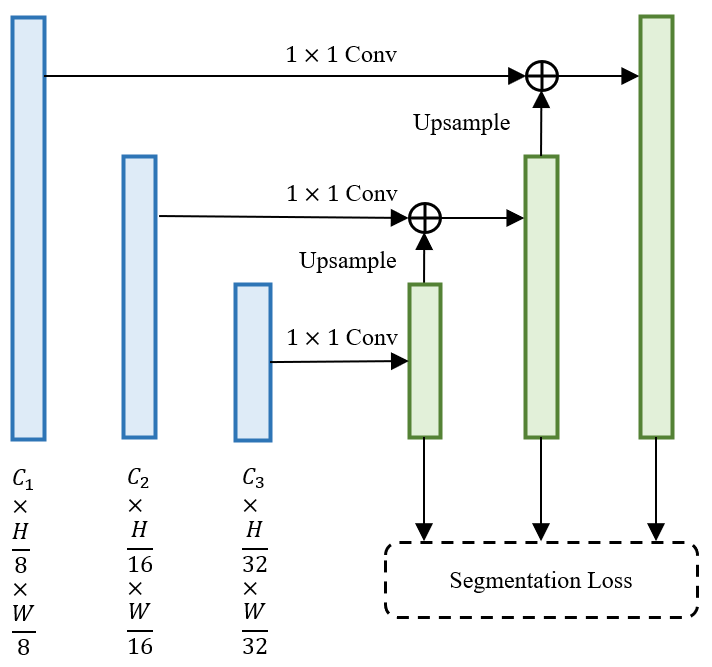}
        \end{subfigure}
    \end{minipage} 

    \caption{Perspective view auxiliary FPN and loss. This structure is only utilized when training and does not influence the inference speed.}
    \vspace{-6mm}
    \label{fig:pv_fpn}
\end{figure}

\begin{figure}[t]
    \centering
    \begin{minipage}{1\linewidth}
        \centering
        \begin{subfigure}{1\linewidth}
        \includegraphics[width=1\textwidth]{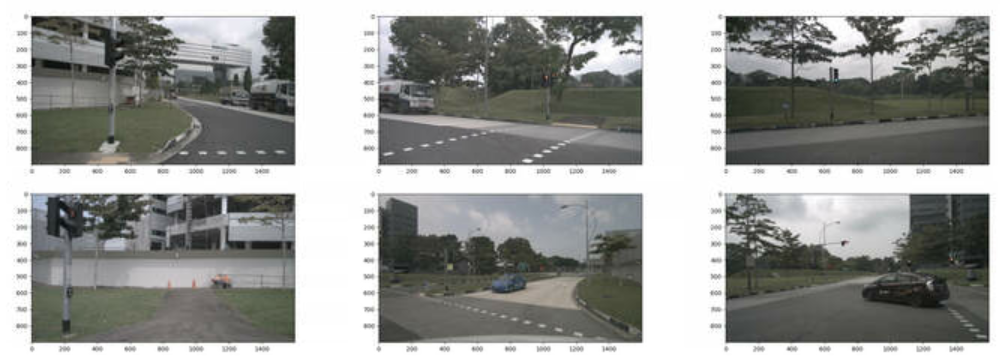}
        \includegraphics[width=1\textwidth]{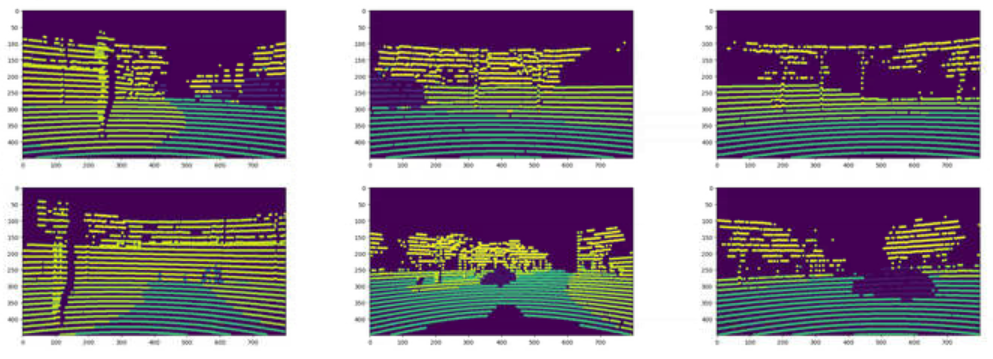}
        \end{subfigure}
    \end{minipage} 
    \vspace{-2mm}
    \caption{Perspective view supervision visualization.}
    \vspace{-6mm}
    \label{fig:pv_fpn_vis}
\end{figure}

%3. PV branch \\
\noindent{\bf{Perspective View Supervision.}}
Due to the presence of the view transformation module such as LSS~\cite{philion2020lift} or BEVFormer~\cite{li2022bevformer}, the perspective image features are first converted into BEV features, which will be passed through a series of complex neural network structures to obtain the final occupancy voxel. As a result, the supervision signals are far from the image backbone, and the learning of parameters in the image backbone may be influenced by the long path of the gradient flow. To mitigate this problem, we introduced a segmentation supervision signal in the perspective view. We found this auxiliary loss can improve performance without additional computational overhead during inference.

As demonstrated in Figure~\ref{fig:pv_fpn}, we attach a U-Net-like network after the image backbone to predict the segmentation mask of each view in the image space. The weights of the plug-in U-Net are shared by each view image. Due to the nuScenes dataset does not provide image-view segmentation ground truth, as demonstrated in Figure~\ref{fig:pv_fpn_vis}, we project the LiDAR points into every view of the image and assign the label of the LiDAR point to the pixel at the corresponding pixel on the image. Consequently, we generate sparse segmentation labels for every view of the image at multi-level features.

We utilize focal loss to supervise the training. Suppose there are $N$ cameras. The level $l$ of the output of the additionally inserted U-Net of each view is represented by $I_{i}^l$. The binary mask $M_{i}^l$ indicates where the LiDAR points are projected. The sparse image-view segmentation ground truth of each view is represented as $S_{i}^l$. The loss $\mathcal{L}_{PV}$ of perspective views can be expressed as,

\begin{equation}
    \mathcal{L}_{PV} = \sum_{i=1}^{N}{\sum_{l=1}^{L}{{\rm FocalLoss} (M_{i}^l * {I_i}^l, S_{i}^l)}}
\end{equation}

% 4. Visible Mask \\
\noindent{\bf{Visible Mask.}}
The generation of 3D occupancy grid ground truth is difficult. Some areas in the scene cannot be observed by onboard sensors, such as behind walls or inside houses. It is blind to predict the occupation of these areas where information cannot be obtained. Therefore, we only consider the visible voxel when training to make the supervision signals more reliable.

\noindent{\bf{Loss Functions.}}
Since there is a large number of empty voxels in the 3D space, we utilize the focal loss~\cite{lin2017focal} for voxel segmentation to balance the number of positive and negative samples. As shown in Equation~\ref{eq:loss_total}, the total loss $\mathcal{L}$ is the weighted summation of the PV loss $\mathcal{L}_{PV}$ and the voxel loss $\mathcal{L}_{vox}$.
\begin{equation}
    \mathcal{L} = \mathcal{W}_{PV}\mathcal{L}_{PV} + \mathcal{W}_{vox}\mathcal{L}_{vox}
    \label{eq:loss_total}
\end{equation}
Here, $\mathcal{W}_{PV}$ and $\mathcal{W}_{vox}$ are represented as the weights of the weights of the PV loss and the voxel loss, respectively.

%5. todo \\
% \subsection{Loss}

\section{Experiments}
\subsection{Dataset Introduction}
\noindent{\bf{OpenOcc.}}
%There are several occupancy benchmarks based on NuScenes Dataset\cite{caesar2020nuscenes}, and they almost generate occupation labels based on point cloud annotations varying in the number of categories and the range of point clouds. 
We choose the OpenOcc ~\cite{sima2023_occnet} to facilitate comparison with the SOTA methods. OpenOcc is generated using dense and high-quality occupancy annotations derived from sparse LiDAR data and 3D bounding boxes based on the nuScenes ~\cite{nuscenes} dataset, which comprises 700 training scenes and 150 validation scenes. The occupancy classes are the same as LiDAR segmentation, including 10 foreground objects and 6 background categories. 

\noindent{\bf{SemanticKITTI.}}
The SemanticKITTI~\cite{behley2019semantickitti} dataset, built upon the widely popular KITTI Odometry dataset, emphasizes on semantic understanding of scenes using LiDAR points and front-view camera image. The semantic occupancy ground truth is annotated with 21 semantic classes (19 semantics, 1 free and 1 unknown). 
% \todo{'感觉这里需要补充一些东西？戛然而止了'}

\subsection{Experiment Setting}
\noindent{\bf{Implementation Details on OpenOcc Dataset.}}
We adopt LiDAR range $[-50m,50m] \times [-50m,50m] \times [-5m,3m]$ and voxel size $[200,200,16]$, to be consistent with OccNet~\cite{sima2023_occnet}. 
Following the experimental setting of BEVFormer~\cite{li2022bevformer} and OccNet, we use two backbones, which are ResNet50 pretrained on ImageNet and ResNet101-DCN initialized from FCOS3D ~\cite{wang2021fcos3d}. 
The ResNet50 output single feature at the size of 1/32, and ResNet101-DCN output multi-scales features from FPN at sizes of 1/8,1/16,1/32.
% For the encoder, the BEV encoder includes 2 types of encoder layers with temporal self-attention and spatial cross-attention. 
We use 3 BEV layers in ResNet50 and 6 BEV layers in ResNet101-DCN respectively. 
The BEV feature dimension is set to 256 and the voxel feature dimension is set to 128.
% In the ResNet101-DCN model, both 3D voxel FPN and efficient voxel FPN exhibit a low latency ratio, with similar overall inference latencies. Consequently, we opted for the 3D voxel FPN for the large backbone ResNet101-DCN and the Partial Voxel FPN for ResNet50 to prioritize efficiency.
We set the learning rate max to $2e^{-4}$, with 0.01 weight decay. The model is trained with 24 epochs. 

\noindent{\bf{Implementation Details on SemanticKITTI Dataset.}}
We adopt LiDAR range $[0,51.2] \times [-25.6m,25.6m] \times [-2m,4.4m]$ and voxel size $[256,256,32]$, to be consistent with OccFormer. We adopt EfficientNet-B7 ~\cite{tan2019efficientnet} as the backbone, following the compared
methods. We generate the 3D feature volume of size $192\times192\times32$ with 128 channels and upsampled to
$256\times256\times32$ for full-scale evaluation. Unless specified, we set the learning rate max to $10^{-4}$, and train the model for 30 epochs.
% 补充semantickitti 的实验设置

\subsection{Metrics}
% The mean intersection over union (mIoU) is used to evaluate the performance of predicted occupancy results on both the semantic scene completion (SSC) and the LiDAR segmentation tasks.
% We also report IoU results for each category.
% The LiDAR segmentation results are derived from the corresponding predicted 3D semantic occupancy volumes, as described in~\cite{sima2023_occnet}.

The mean intersection over union (mIoU) metric is employed to evaluate the performance of the predicted occupancy results for both the semantic scene completion (SSC) and the LiDAR segmentation tasks. We also provide intersection over union (IoU) scores for individual categories. The LiDAR segmentation results are obtained from the correspondingly predicted 3D semantic occupancy volumes, as detailed in~\cite{sima2023_occnet}.

% \todo{EB: can we provide more detail on the why? How many parameters or the compute of each model?}.

\subsection{Comparison with State-of-the-Art Methods}
\noindent{\bf{Occupancy Prediction on OpenOcc Dataset.}}
Table~\ref{tab:occ_openocc_val} shows the results comparing our method with previous SOTA methods for occupancy tasks.
We report the results of BEVDet4D~\cite{bevdet}, BEVDepth~\cite{bevdepth} and BEVDet~\cite{bevdet} reproduced by OccNet~\cite{sima2023_occnet} via replacing the detection head with the scene completion head built on their BEV feature maps on ResNet50 backbone.
Compared with the TPVFormer~\cite{tpvformer}, our method has a substantial boost of 3.55\% mIoU on Res101-DCN backbone model. 
As for the SOTA method OccNet on OpenOcc dataset, we compared it on both ResNet50 and ResNet101-DCN. Our method surpasses OcNet by 1.64\% with about $1/3$ inference latency on ResNet50 setting, and also obtains a little higher results on Res101-DCN with fewer inference latency shown in Table~\ref{tab:speed}.
% Methods with * stand for training and evaluating on OpenOcc dataset reproduced by OccNet~\cite{sima2023_occnet}.
% Methods with $^{+}$ stand for our reproduced results.

% occ result table
\begin{table*}
% \vspace{-6mm}
\centering
% \begin{tabular}{c|c|*{2}{p{0.6cm}<{\centering}}|*{17}{p{0.5cm}<{\centering}}}
\resizebox{\linewidth}{!}{
\begin{tabular}{l|c|*{2}{c}|*{16}{c}}
\toprule  %添加表格头部粗线
Method & Backbone &  IoU  &  mIoU  & \begin{turn}{90}{Car}\end{turn} & \begin{turn}{90}{Truck}\end{turn} & \begin{turn}{90}{Trailer}\end{turn} & \begin{turn}{90}{Bus}\end{turn} & \begin{turn}{90}{const. veh.}\end{turn} & \begin{turn}{90}{Bicycle}\end{turn} & \begin{turn}{90}{Motorcycle}\end{turn} & \begin{turn}{90}{Pedestrian}\end{turn} & \begin{turn}{90}{Traffic cone}\end{turn} & \begin{turn}{90}{Barrier}\end{turn} & \begin{turn}{90}{Drivable surface}\end{turn} & \begin{turn}{90}{Other flat}\end{turn} & \begin{turn}{90}{Sidewalk}\end{turn} & \begin{turn}{90}{Terrain}\end{turn} & \begin{turn}{90}{Manmade}\end{turn} & \begin{turn}{90}{Vegetation}\end{turn} \\
\midrule  %添加表格中横线
BEVDet4D$^{\dagger}$~\cite{huang2022bevdet4d} & ResNet50 & 18.27 & 9.85 & 26.98 & 12.63 & 1.93 & 13.04 & 0.61 & 0.00 & 1.20 & 6.76 & 0.93 & 13.56 & 27.23 & 11.09 & 13.64 & 12.04 & 6.42 & 9.56 \\
BEVDepth$^{\dagger}$~\cite{bevdepth} & ResNet50 & 23.45 & 11.88 & 27.05 & 14.31 & 1.91 & 20.75 & 1.10 & 0.02 & 2.01 & 9.69 & 1.45 & 15.15 & 31.92 & 7.88 & 17.08 & 16.27 & 8.76 & 14.75 \\
BEVDet$^{\dagger}$~\cite{bevdet} & ResNet50 & 27.46 & 12.49 & 21.09 & 13.76 & 3.4 & 18.27 & 2.62 & 0.11 & 1.42 & 7.78 & 1.08 & 16.06 & 33.89 & 10.84 & 17.55 & 22.03 & 11.72 & 18.15 \\
BEVNet$^{\dagger}$~\cite{sima2023_occnet} & ResNet50 & 36.11 & 17.37 & 24.94 & 16.02 & 10.21 & 20.85 & 8.64 & 5.07 & 7.75 & 12.80 & 8.93 & 14.02 & 44.41 & 14.42 & 23.87 & 27.76 & 13.73 & 24.49 \\
BEVNet*~\cite{sima2023_occnet} & ResNet50 & 36.41 & 18.45 & 26.04 & 17.53 & 9.34 & 21.76 & 8.98 & 5.72 & 8.67 & 12.31 & 7.87 & 19.21 & 45.21 & 15.52 & 26.27 & 28.92 & 15.83 & 26.01\\
OccNet~\cite{sima2023_occnet} & ResNet50 & 37.69 & 19.48 & \textbf{27.72} & 18.00 & 10.68 & 24.16 & 9.79 & 5.52 & 7.73 & \textbf{13.38} & 7.18 & \textbf{20.63} & 46.13 & 20.60 & 26.75 & 29.37 & 16.90 & 27.21\\
% Ours & ResNet50 & 39.21 & 21.01 & 27.49 & 19.88 & 13.10 & 25.17 & 10.82 & 8.25 & 10.38 & 13.55 & 8.95 & 21.92 & 47.57 & 23.87 & 28.81 & 31.12 & 17.93 & 27.42 \\
% Ours & ResNet50 & 39.03 & \textcolor{red}{20.98} & 26.93 & 19.61 & 12.18 & 24.45 & 11.79 & 8.69 & 11.56 & 13.17 & 10.98 & 19.80 & 47.31 & 25.10 & 28.67 & 30.83 & 17.16 & 27.48 \\
Ours & ResNet50 & \textbf{39.29} & \textbf{21.12} & 27.43 & \textbf{19.79} & \textbf{12.75} & \textbf{24.82} & \textbf{11.79} & \textbf{8.65} & \textbf{11.53} & 13.22 & \textbf{10.94} & 20.24 & \textbf{47.33} & \textbf{25.25} & \textbf{28.54} & \textbf{30.82} & \textbf{17.35} & \textbf{27.52} \\
\midrule
TPVFormer$^{\dagger}$~\cite{tpvformer} & Res101-DCN & 37.47 & 23.67 & \textbf{38.70} & 26.47 & 16.42 & 33.24 & 12.41 & 12.75 & 17.84 & 11.65 & 8.49 & 12.75 & 47.88 & 25.43 & 30.62 & 30.18 & 15.51 & 23.12 \\
BEVNet$^{\dagger}$~\cite{sima2023_occnet} & Res101-DCN & 40.15 & 24.62 & 35.83 & 23.90 & 12.82 & 32.07 & 11.93 & 15.79 & 19.72 & 19.75 & 15.38 & 26.39 & 46.13 & 20.60 & 26.75 & 29.37 & 16.90 & 27.21 \\
OccNet~\cite{sima2023_occnet} & Res101-DCN & 41.08 & 26.98 & 37.35 & 26.23 & 16.37 & 34.16 & \textbf{15.58} & 16.89 & \textbf{21.92} & \textbf{21.29} & \textbf{16.75} & \textbf{29.77} & 50.74 & 27.93 & 31.98 & 33.24 & 20.80 & 30.68 \\
Ours & Res101-DCN & \textbf{43.06} & \textbf{27.22} & 36.16 & \textbf{26.53} & \textbf{17.09} & \textbf{34.52} & 15.42 & \textbf{17.06} & 21.41 & 19.61 & 16.67 & 28.88 & \textbf{50.95} & \textbf{28.69} & \textbf{33.49} & \textbf{35.10} & \textbf{22.32} & \textbf{31.66}   \\ 

% 顺序需要调整，两个paper的类别顺序不一致
\bottomrule %添加表格底部粗线
\end{tabular}
}
% \caption{Semantic scene completion performance on OpenOcc~\cite{sima2023_occnet} validation set. Methods with $^{\dagger}$ stand for training and evaluating on OpenOcc dataset reproduced by OccNet~\cite{sima2023_occnet}, and methods with * stand for our reproducing.}
\caption{3D occupancy prediction in terms of semantic scene completion performance on OpenOcc~\cite{sima2023_occnet} validation set. Methods with $^{\dagger}$ stand for training and evaluating on OpenOcc dataset reproduced by OccNet~\cite{sima2023_occnet}, and methods with * stand for our reproducing.}
\vspace{-8mm}
\label{tab:occ_openocc_val}
\end{table*}

\noindent{\bf{Occupancy Prediction for LiDAR Segmentation.}}
Following the practices from OccFormer~\cite{occformer}, the LiDAR semantic segmentation task is utilized as a quantitative indicator for the 3D semantic occupancy prediction. We adopt the same strategy as OccNet to transfer semantic occupancy prediction to LiDAR segmentation by assigning the
point label based on the associated voxel label without LiDAR supervision.
As shown in Table~\ref{tab:lidarseg_nus_val}, %our method
%outperforms the only vision-based method OccNet and achieves comparable performance with the state-of-the-art LiDAR-based methods.
our method surpasses OccNet by 7.92\% with ResNet50 backbone, and significantly ahead of all foreground classes except pedestrian. Similarly, our method surpasses OccNet by 1.36\% with ResNet101 backbone while requiring fewer inference latency shown in Table~\ref{tab:speed}.

% lidar seg table

\begin{table*}[htbp]
\vspace{-6mm}
\centering
% \begin{tabular}{c|c|*{2}{p{0.6cm}<{\centering}}|*{17}{p{0.5cm}<{\centering}}}
\resizebox{\linewidth}{!}{
\begin{tabular}{l|c|c|*{16}{c}}
\toprule  %添加表格头部粗线
Method & Backbone & mIoU  & \begin{turn}{90}{Car}\end{turn} & \begin{turn}{90}{Truck}\end{turn} & \begin{turn}{90}{Trailer}\end{turn} & \begin{turn}{90}{Bus}\end{turn} & \begin{turn}{90}{const. veh.}\end{turn} & \begin{turn}{90}{Bicycle}\end{turn} & \begin{turn}{90}{Motorcycle}\end{turn} & \begin{turn}{90}{Pedestrian}\end{turn} & \begin{turn}{90}{Traffic cone}\end{turn} & \begin{turn}{90}{Barrier}\end{turn} & \begin{turn}{90}{Drivable surface}\end{turn} & \begin{turn}{90}{Other flat}\end{turn} & \begin{turn}{90}{Sidewalk}\end{turn} & \begin{turn}{90}{Terrain}\end{turn} & \begin{turn}{90}{Manmade}\end{turn} & \begin{turn}{90}{Vegetation}\end{turn} \\
\midrule  %添加表格中横线

OccNet~\cite{sima2023_occnet} & ResNet50 & 47.29 & 63.05 & 41.12 & 23.38 & 48.32 & 24.12 & 20.64 & 20.24 & \textbf{41.82} & 18.84 & 59.06 & \textbf{86.46} & 53.12 & 52.03 & 59.14 & \textbf{71.55} & \textbf{73.68} \\

Ours & ResNet50 & \textbf{55.21} & \textbf{66.82} & \textbf{65.77} & \textbf{52.26} & \textbf{81.16} & \textbf{34.91} & \textbf{23.21} & \textbf{35.57} & 37.46 & \textbf{23.97} & \textbf{60.90} & 85.93 & \textbf{60.90} & \textbf{58.34} & \textbf{62.26} & 65.19 & 68.71 \\

\midrule
TPVFormer$^{\dagger}$~\cite{tpvformer} & Res101-DCN & 58.45 & \textbf{74.28} & 70.79 & 53.96 & 80.88 & \textbf{47.04} & 24.50 & 47.09 & 33.42 & 14.52 & 65.99 & 88.55 & 61.63 & 59.46 & 63.15 & 75.76 & 74.17 \\

OccNet~\cite{sima2023_occnet} & Res101-DCN & 60.46 & 73.88 & 67.08 & 52.20 & 77.37 & 37.62 & \textbf{32.58} & 50.87 & \textbf{51.45} & 33.69 & \textbf{66.95} & 88.72 & 57.99 & 58.04 & 63.06 & \textbf{78.91} & \textbf{76.97} \\

Ours & Res101-DCN & \textbf{61.82} & 70.86 & \textbf{71.23} & \textbf{55.89} & \textbf{83.60} & 43.74 & 31.12 & \textbf{53.15} & 48.41 & \textbf{34.36}  & 65.89 & \textbf{89.07} & \textbf{64.62} & \textbf{63.75} & \textbf{65.50} & 75.13 & 73.81 \\

\bottomrule %添加表格底部粗线
\end{tabular}
}
\caption{LiDAR segmentation performance on nuScenes~\cite{nuscenes} validation set.}
\vspace{-8mm}
\label{tab:lidarseg_nus_val}
\end{table*}

\noindent{\bf{Semantic Scene Completion on SemanticKITTI Dataset.}}
Table~\ref{tab:semantickitti_result} compares our method with existing occupancy network methods for the semantic scene completion task on the SemanticKITTI test set. The mIoU metric of our method surpasses other SOTA methods like TPVFormer~\cite{tpvformer} by 1.18\% and OccFormer~\cite{occformer} by 0.16\%. Especially, as a BEV-based method, our model is good at detecting ground-like elements like 'parking' and 'other ground'. The iou metric of our method about the 'Other grounding' category surpasses the OccFormer by 4.9\%.
% We compare our Fast Occupancy against TPVFormer~\cite{tpvformer}, which is a
% vision-based method. 

% Compared with the recent OccFormer~\cite{occformer}, our method achieves
% a remarkable boost of? mIoU.
% The results demonstrate the effectiveness of our method in different types of datasets.

% semantic kitti test
\begin{table*}
\centering
% \begin{tabular}{c|c|*{2}{p{0.6cm}<{\centering}}|*{17}{p{0.5cm}<{\centering}}}

\resizebox{\linewidth}{!}{
\begin{tabular}{c|c|c|*{19}{c}}
\toprule  %添加表格头部粗线
% Method  & \makecell{IoU \\ (\%)} & \begin{turn}{90}{Road}\end{turn} & \begin{turn}{90}{Sidewalk}\end{turn} & \begin{turn}{90}{Parking}\end{turn} & \begin{turn}{90}{Other ground}\end{turn} & \begin{turn}{90}{Building}\end{turn} & \begin{turn}{90}{Car}\end{turn} & \begin{turn}{90}{Truck}\end{turn} & \begin{turn}{90}{Bicycle}\end{turn} & \begin{turn}{90}{Motorcycle}\end{turn} & \begin{turn}{90}{Other Vehicle}\end{turn} & \begin{turn}{90}{Vegetation}\end{turn} & \begin{turn}{90}{Trunk}\end{turn} & \begin{turn}{90}{Terrain}\end{turn} & \begin{turn}{90}{Person}\end{turn} & \begin{turn}{90}{Bicyclist}\end{turn} & \begin{turn}{90}{Motorcyclist}\end{turn} & \begin{turn}{90}{Fence}\end{turn} & \begin{turn}{90}{Pole}\end{turn} & \begin{turn}{90}{Traffic sign}\end{turn} & \makecell{mIoU \\ (\%)}\\
Method  & mIoU & IoU  & \begin{turn}{90}{Road}\end{turn} & \begin{turn}{90}{Sidewalk}\end{turn} & \begin{turn}{90}{Parking}\end{turn} & \begin{turn}{90}{Other ground}\end{turn} & \begin{turn}{90}{Building}\end{turn} & \begin{turn}{90}{Car}\end{turn} & \begin{turn}{90}{Truck}\end{turn} & \begin{turn}{90}{Bicycle}\end{turn} & \begin{turn}{90}{Motorcycle}\end{turn} & \begin{turn}{90}{Other Vehicle}\end{turn} & \begin{turn}{90}{Vegetation}\end{turn} & \begin{turn}{90}{Trunk}\end{turn} & \begin{turn}{90}{Terrain}\end{turn} & \begin{turn}{90}{Person}\end{turn} & \begin{turn}{90}{Bicyclist}\end{turn} & \begin{turn}{90}{Motorcyclist}\end{turn} & \begin{turn}{90}{Fence}\end{turn} & \begin{turn}{90}{Pole}\end{turn} & \begin{turn}{90}{Traffic sign}\end{turn}\\
\midrule  %添加表格中横线

% LMSCNet*~\cite{roldao2020lmscnet} & 7.07 & 31.37 & 46.7 & 19.5 & 13.5 & 3.1 & 10.3 & 14.3 & 0.3 & 0.0 & 0.0 & 0.0 & 10.8 & 0.0 & 10.4 & 0.0 & 0.0 & 0.0 & 5.4 & 0.0 & 0.0 \\
% 3DSketch*~\cite{chen20203d}  & 6.23 & 26.85 & 37.7 & 19.8 & 0.0 & 0.0 & 12.1 & 17.1 & 0.0 &0.0 & 0.0 & 0.0& 12.1 & 0.0 & 16.1 & 0.0 & 0.0 & 0.0 & 3.4 & 0.0 & 0.0 \\
% AICNet*~\cite{li2020anisotropic} & 7.09 & 23.93 & 39.3 & 18.3 & 19.8 & 1.6 & 9.6 & 15.3 & 0.7 & 0.0 & 0.0 & 0.0 & 9.6 & 1.9 & 13.5 & 0.0 & 0.0 & 0.0 & 5.0 & 0.1 & 0.0 \\
% JS3C-Net*~\cite{yan2021sparse} & 8.97 & 34.00 & 47.3 & 21.7 & 19.9 & 2.8 & 12.7 & 20.1 & 0.8 & 0.0 & 0.0 & 4.1 & 14.2 & 3.1 & 12.4 & 0.0 & 0.2 & 0.2 & 8.7 & 1.9 & 0.3 \\
% MonoScene†~\cite{monoscene} & 11.08 & 34.16 & 54.7 & 27.1 & 24.8 & 5.7 & 14.4 & 18.8 & 3.3 & 0.5 & 0.7 & 4.4 & 14.9 & 2.4 & 19.5 & 1.0 & 1.4 & 0.4 & 11.1 & 3.3 & 2.1\\
% VoxFormer~\cite{voxformer} & 12.20 & 42.95 & 53.9 & 25.3 & 21.1 & 3.7 & 19.8 & 20.8 & 3.5 & 1.0 & 0.7 & 3.7 & 22.4 & 7.5 & 21.3 & 1.4 & 2.6 & 0.2 & 11.1 & 5.1 & 4.9 \\
% TPVFormer~\cite{tpvformer} & 11.26 & 34.25 & 55.1 & 27.2 & 27.4 & 6.5 & 14.8 & 19.2 & 3.7 & 1.0 & 0.5 & 2.3 & 13.9 & 2.6 & 20.4 & 1.1 & 2.4 & 0.3 & 11.0 & 2.9 & 1.5 \\
% OccFormer~\cite{occformer} & 12.32 & 34.53 & 55.9 & 30.3 & 31.5 & 6.5 & 15.7 & 21.6 & 1.2 & 1.5 & 1.7 & 3.2 & 16.8 & 3.9 & 21.3 & 2.2 & 1.1 & 0.2 & 11.9 & 3.8 & 3.7 \\
% \midrule  %添加表格中横线
% Ours & \textbf{12.44} & 33.40 & 55.9 & 29.6 & 32.8 & 11.4 & 15.0 & 21.2 & 2.2 & 1.1 & 1.6 & 4.4 & 15.9 & 2.2 & 19.8 & 1.3 & 1.1 & 0.5 & 11.7 & 3.3 & 2.1 \\ 
LMSCNet*~\cite{roldao2020lmscnet} & 7.07 & 31.37 & 46.70 & 19.50 & 13.50 & 3.10 & 10.30 & 14.30 & 0.30 & 0.00 & 0.00 & 0.00 & 10.80 & 0.00 & 10.40 & 0.00 & 0.00 & 0.00 & 5.40 & 0.00 & 0.00 \\
3DSketch*~\cite{chen20203d}  & 6.23 & 26.85 & 37.70 & 19.80 & 0.00 & 0.00 & 12.10 & 17.10 & 0.00 &0.00 & 0.00 & 0.00& 12.10 & 0.00 & 16.10 & 0.00 & 0.00 & 0.00 & 3.40 & 0.00 & 0.00 \\
AICNet*~\cite{li2020anisotropic} & 7.09 & 23.93 & 39.30 & 18.30 & 19.80 & 1.60 & 9.60 & 15.30 & 0.70 & 0.00 & 0.00 & 0.00 & 9.60 & 1.90 & 13.50 & 0.00 & 0.00 & 0.00 & 5.00 & 0.10 & 0.00 \\
JS3C-Net*~\cite{yan2021sparse} & 8.97 & 34.00 & 47.30 & 21.70 & 19.90 & 2.80 & 12.70 & 20.10 & 0.80 & 0.00 & 0.00 & 4.10 & 14.20 & 3.10 & 12.40 & 0.00 & 0.20 & 0.20 & 8.70 & 1.90 & 0.30 \\
MonoScene~\cite{monoscene} & 11.08 & 34.16 & 54.70 & 27.10 & 24.80 & 5.70 & 14.40 & 18.80 & 3.30 & 0.50 & 0.70 & 4.40 & 14.90 & 2.40 & 19.50 & 1.00 & 1.40 & 0.40 & 11.10 & 3.30 & 2.10\\
VoxFormer~\cite{voxformer} & 12.20 & 42.95 & 53.90 & 25.30 & 21.10 & 3.70 & 19.80 & 20.80 & 3.50 & 1.00 & 0.70 & 3.70 & 22.40 & 7.50 & 21.30 & 1.40 & 2.60 & 0.20 & 11.10 & 5.10 & 4.90 \\
TPVFormer~\cite{tpvformer} & 11.26 & 34.25 & 55.10 & 27.20 & 27.40 & 6.50 & 14.80 & 19.20 & \textbf{3.70} & 1.00 & 0.50 & 2.30 & 13.90 & 2.60 & 20.40 & 1.10 & \textbf{2.40} & 0.30 & 11.00 & 2.90 & 1.50 \\
OccFormer~\cite{occformer} & 12.32 & \textbf{34.53} & 55.90 & \textbf{30.30} & 31.50 & 6.50 & \textbf{15.70} & \textbf{21.60} & 1.20 & \textbf{1.50} & \textbf{1.70} & 3.20 & \textbf{16.80} & \textbf{3.90} & \textbf{21.30} & \textbf{2.20} & 1.10 & 0.20 & \textbf{11.90} & \textbf{3.80} & \textbf{3.70} \\
\midrule  %添加表格中横线
Ours & \textbf{12.44} & 33.40 & \textbf{55.90} & 29.60 & \textbf{32.80} & \textbf{11.40} & 15.00 & 21.20 & 2.20 & 1.10 & 1.60 & \textbf{4.40} & 15.90 & 2.20 & 19.80 & 1.30 & 1.10 & \textbf{0.50} & 11.70 & 3.30 & 2.10 \\

\bottomrule %添加表格底部粗线
\end{tabular}
}
\caption{ Semantic scene completion results on SemanticKITTI~\cite{behley2019semantickitti} test set. * represents these methods are implemented by MonoScene~\cite{monoscene}.}
\vspace{-8mm}
\label{tab:semantickitti_result}
\end{table*}

\subsection{Speed Comparison}
% We prioritize inference latency as the evaluation criterion of computational efficiency instead of parameters and FLOPs, as it provides a more direct measure. As for the number of parameters and flops, while important, does not have a direct correlation with latency.
We prioritize inference latency as the evaluation criterion for computational efficiency over parameters and FLOPs because it offers a more direct reflection of performance.
We compare our method with state-of-the-art occupancy methods in terms of relative inference latency on AMD MI100 and MI250 GPUs. We use BEVNet-R50 as a baseline to benchmark the latency ratio for each GPU respectively.
As shown in Table~\ref{tab:speed}, our method can surpass OccNet by about 1.7\% with close to 3$\times$ (1.22$\times$ vs 3.43$\times$) speedups on ResNet50 backbone model and slightly higher results with 1.47$\times$ faster (3.88$\times$ vs 5.71$\times$) on ResNet101-DCN backbone model. 
Compared with BEVNet as a baseline, our method obtains 2.4\% gain by only increasing the latency by 10.6\% (3.88$\times$ vs 3.63$\times$) on ResNet101-DCN backbone model, and 3.75\%  gain with 22\%(1.22$\times$ vs 1.00$\times$) extra time-cost on ResNet50 backbone model respectively. 
In the SemanticKITTI benchmark, we compare our method with OccFormer~\cite{occformer}. Even if the number of parameters and latency of our method are about half that of OccFormer, the accuracy of our method is also higher than OccFomer.
Overall, our method can obtain the best performance in terms of mIOU and IoU with efficiency and effectiveness.
% \todo{这部分缺少对SemanticKITTI数据几下EfficientNet B7的对比描述}

\begin{table*}[htbp]
\vspace{-6mm}
\centering
% \begin{tabular}{c|c|*{2}{p{0.6cm}<{\centering}}|*{17}{p{0.5cm}<{\centering}}}

% \resizebox{\linewidth}{!}{
\scalebox{0.8}{
\begin{tabular}{l|c|c|c|c|c|c}
\toprule  %添加表格头部粗线

Methods & Benchmark & Input size & Backbone &  mIOU   & \makecell{Latency$\downarrow$ \\ (MI100)}  & \makecell{Latency$\downarrow$ \\ (MI250)} \\
\midrule  %添加表格中横线
BEVNet~\cite{sima2023_occnet} & OpenOcc & 450x800 & ResNet50 & 17.37  & $1.00\times$ & $1.00\times$ \\
OccNet~\cite{sima2023_occnet} & OpenOcc   & 450x800 & ResNet50 &19.48  & 3.30$\times$ & 3.43$\times$ \\

Ours  & OpenOcc & 450x800 & ResNet50 & \textbf{21.12}  & 1.22$\times$ & 1.22$\times$\\

\midrule  %添加表格中横线

BEVNet~\cite{sima2023_occnet} & OpenOcc & 900x1600 & ResNet101-DCN & 24.62  & 4.04$\times$ & 3.63$\times$\\

TPVFormer~\cite{tpvformer} & OpenOcc  & 900x1600 & ResNet101-DCN & 23.67  & 5.05$\times$ & 4.68$\times$ \\
OccNet~\cite{sima2023_occnet} & OpenOcc  & 900x1600 & ResNet101-DCN & 26.98  & 6.28$\times$ & 5.71$\times$\\
Ours & OpenOcc  & 900x1600 & ResNet101-DCN & \textbf{27.02} & 4.31$\times$ & 3.88$\times$ \\

\midrule  %添加表格中横线

% TPVFormer ~\cite{tpvformer}  & \textcolor{red}{928x1600} ? & EfficientNet B7 &-&-& - & 642ms & 648ms\\
VoxFormer~\cite{voxformer} & SemanticKITTI  & 370x1220 & ResNet50 & 12.20  & 4.02$\times$ & 3.61$\times$\\
OccFormer~\cite{occformer} & SemanticKITTI  & 256x704 & EfficientNet B7 & 12.32  & 5.06$\times$ & 4.61$\times$\\

Ours  & SemanticKITTI  & 256x704 & EfficientNet B7 & \textbf{12.44}  &2.90$\times$ & 2.44$\times$ \\ %(192x192)

\bottomrule %添加表格底部粗线
\end{tabular}
}
\caption{Efficiency and performance analysis with model structure.}
\vspace{-12mm}
\label{tab:speed}
\end{table*}

\subsection{Ablation Studies}
\noindent{\bf{Effect of Our Lifting BEV Feature Method.}}
\label{sec_ablation_uplift}
We investigate the effect of lifting methods by comparing the mIOU and latency of different methods. To facilitate an intuitive sense of the module's size, we also present the proportion of latency in the overall. As shown in Table~\ref{tab:ablation_uplift}, we utilize BEVNet as the baseline, which is BEVFormer + MLP. We observe that the naive method of employing MLP to expand the channel number and reshape it to the height dimension yields relatively poor performance. The main reason is that the method does not take into account the neighboring pixels, resulting in a receptive field of only 1. This is why we have a 0.7\% gain by simply replacing MLP with Conv2D.
% We get a 0.8\% gain from continuing to expand the receptive field with dilation Conv2D, but we don't gain more from increasing the dilation.
We continue to expand the receptive field by using a $5\times5 $ kernel of convolution, but the improvement is minimal and comes with significant latency.
The Deformable Conv2D can continue to increase the receptive field while having the same dynamically irregular reference points, and achieve 0.9\% improvement compared to Conv2D. 
The deformable offset in Deformable Conv2D is necessary because the shapes of objects at different heights are not consistent.
% We also compare our method with other methods to obtain 3D features. 
% The deformable conv2D can achieve comparable performance but 3x faster than Conv3D.
We also utilize the 3D Deformable Attention proposed by OccNet and directly lift the BEV feature, the performance is close to the original OccNet but lower than our method.
% Specifically, 3D Deformable Cross Attention reshapes the BEV feature as query and attention with multi-view image features which is not getting attention feature from the BEV feature. It can obtain the information from the multi-view image features very well, but it is not a good lifting method.
\begin{table}[h!]
\vspace{-6mm}
\centering
% \begin{tabular}{c|c|*{2}{p{0.6cm}<{\centering}}|*{17}{p{0.5cm}<{\centering}}}
\scalebox{0.8}{
% \resizebox{\linewidth}{!}{
\begin{tabular}{l|c|c|c|c}
\toprule  %添加表格头部粗线
% Methods of Uplift BEV Feature & mIOU &  IoU$_{geo}$ & speed  \\
Methods & mIOU &  IoU & \makecell{Latency$\downarrow$ \\ (MI250)} & \makecell{Proportion \\ of Overall} \\
\midrule  %添加表格中横线

% BEVFormer + MLP (Baseline) & 17.37 & 36.11 & 2ms & 2.1\% \\
BEVFormer + MLP (Baseline)* & 18.45 & 36.41 & 1.0$\times$ & 2.1\% \\
BEVFormer + Conv2D & 19.14 & 37.57 & 3.5$\times$ & 7.5\%  \\
BEVFormer + Conv2D-5$\times$5 & 19.24 & 36.99 & 13.5$\times$ & 24.5\%  \\
% BEVFormer + Conv2D + dilation & 19.94 & 38.11 & 29ms  \\
%BEVFormer + Deformable Conv2d & 20.03 & 38.21 & 34ms  \\
% BEVFormer + MLP + Conv3D & \textcolor{red}{TBD} & \textcolor{red}{TBD} & \textcolor{red}{TBD}  \\
% BEVFormer + Stacked 3D Deform Attns~\cite{sima2023_occnet} & 19.48 & 37.69 & 142ms  \\
% BEVFormer +  3D Deform Attn & 19.56 & 37.00 & 142ms  \\
BEVFormer +  3D Deform Attn & 19.56 & 37.00 & 48.0$\times$ & 50.7\% \\
BEVFormer + Deformable Conv2D & \textbf{20.03} & \textbf{38.21} & 5.4$\times$ & 13.1\%  \\
\bottomrule %添加表格底部粗线
\end{tabular}
}
\caption{Ablation on different lifting BEV feature methods on ResNet50 backbone model. Methods with * stands for our reproducing.}
\vspace{-8mm}
\label{tab:ablation_uplift}
\end{table}

\noindent{\bf{Effect of Partial Voxel FPN.}}
\label{sec_ablation_pvf}
To identify a more efficient voxel FPN method, we compare several voxel FPN methods while using BEVNet as the baseline without voxel FPN. We reproduce the method from FB-OCC~\cite{li2023fbocc}, which involves downsampling through 3D convolution, upsampling features at different scales, and summing them with the original scale features. Although this method is effective, it is quite time-consuming, accounting for 26.2\% of the overall model latency.
% We also compare our Partial Voxel FPN with the full Conv2D method, the Partial Voxel FPN can achieve almost performance as well as faster inference time.
% Due to the presence of $1\times1$ Conv2D, the original preserved feature can also be adapted to multi-scale features, which is what makes it fast and good.
In our comparison of Partial Voxel FPN with the full Conv2D method, we find that Partial Voxel FPN can achieve nearly equivalent performance while offering faster inference times. The $1\times1$ Conv2D enables the original preserved features to be efficiently adapted to multi-scale representations, contributing to the speed and effectiveness of our approach.
% We also verify the effect of little 3D convolution on the smallest scale, our method is 0.3\% higher than no little 3D convolution.
We have also evaluated the impact of using a small 3D convolution on the smallest scale feature of $50\times50\times4$. Our method demonstrates a 0.3\% improvement in performance compared to not using a small 3D convolution.
Overall, our proposed method can achieve comparable mIOU with a 0.24$\times$ latency of Voxel FPN with full Conv3D.

\begin{table}[htbp]
\centering
\scalebox{0.8}{
% \begin{tabular}{c|c|*{2}{p{0.6cm}<{\centering}}|*{17}{p{0.5cm}<{\centering}}}
% \resizebox{\linewidth}{!}{
\begin{tabular}{l|c|c|c|c}
\toprule  %添加表格头部粗线
% Methods of Uplift BEV Feature & mIOU &  IoU$_{geo}$ & speed  \\
Methods & mIOU &  IoU & \makecell{Latency$\downarrow$ \\ (MI250)} & \makecell{Proportion \\ of Overall}  \\
\midrule  %添加表格中横线
% Baseline & 17.37 & 36.11 & None & None \\
% Baseline* & 18.45 & 36.41 & None & None\\
No Voxel FPN(Baseline)* & 18.45 & 36.41 & None & None\\
% Voxel FPN with Full Conv3D & 19.99 & 37.86 & 65ms &   \\
Voxel FPN with Full Conv3D & 19.99 & 37.86 & 1.00$\times$ & 26.2\%  \\
%Full Voxel FPN with Conv2D & \textcolor{red}{19.58} & 38.50 & 40ms  \\
% Full Voxel FPN with Conv2D & 19.58 & 38.50 & 40ms  \\
Full Voxel FPN with Conv2D & 19.64 & 38.68 & 0.63$\times$ & 20.8\% \\
Partial Voxel FPN w/o Little Conv3D & 19.51 & 37.94 & 0.21$\times$ & 8.4\%  \\
Partial Voxel FPN & 19.75 & 38.13 & 0.24$\times$ & 9.6\%  \\
\bottomrule %添加表格底部粗线
\end{tabular}
% }
}
\caption{Ablation on inference latency of various Voxel FPNs on ResNet50 backbone model. Methods with * stands for our reproducing.}
% \caption{Ablation on inference latency of various Voxel FPNs on ResNet50 backbone model.}
\vspace{-8mm}
\label{tab:ablation_voxelfpn}
% \vspace{-4mm}

\end{table}

\noindent{\bf{Contribution of Our Proposed Methods.}}
To clearly demonstrate the effectiveness of our proposed methods, we conduct ablation experiments as presented in Table ~\ref{tab:ablation_contrib_module}. 
We verify the effectiveness of our methods by comparing them to our reproduced baseline, which is BEVNet, respectively. As mentioned before, we finally pick up the deformable convolution lifting method, which is able to bring a 1.58\% improvement. We also verify the addition of segmentation supervision to the perspective view image features, which brings a 1.14\% boost for free. Visible mask in training is effective in reducing the perturbation of blind areas, which brings a 0.46\% boost. Our proposed Partial Voxel Fpn module can efficiently provide multi-scaled features, and bring 1.30\% improvement. Finally, we converge all the proposed methods and obtain 21.12\% mIOU.

\begin{table}[htbp]
\vspace{-6mm}
\centering
% \begin{tabular}{c|c|*{2}{p{0.6cm}<{\centering}}|*{17}{p{0.5cm}<{\centering}}}
\scalebox{0.8}{
% \resizebox{\linewidth}{!}{
\begin{tabular}{l|cccc|c|c}
\toprule  %添加表格头部粗线
Methods & \makecell{Deformable\\Conv2D} & \makecell{PV\\Supervision} & \makecell{Visible\\Mask}& \makecell{Partial\\Voxel FPN} & mIOU & IoU   \\
% Methods & Deformable Conv2D & PV Supervision & Visible Mask& Partial Voxel FPN & mIOU & IoU   \\
\midrule  %添加表格中横线
% Baseline & & & & & 17.37 & 36.11  \\
Baseline* & & & & & 18.45 & 36.41  \\
Baseline* & \checkmark & & & & 20.03 & 38.21  \\
Baseline*& &\checkmark & & & 19.59 &  37.73  \\
Baseline*& & & \checkmark& & 18.91 & 37.15 \\
Baseline*& & & &\checkmark & 19.75 & 38.13   \\
Baseline*&\checkmark & & &\checkmark & 20.43 & 38.67   \\
% & \checkmark &\checkmark&\checkmark& \checkmark& 21.01 & 39.21   \\
Baseline*& \checkmark &\checkmark&\checkmark& \checkmark& \textbf{21.12} & \textbf{39.29}   \\

\bottomrule %添加表格底部粗线
\end{tabular}
}
\caption{Ablation on contributions of the proposed methods on ResNet50 backbone model. Methods with * stands for our reproducing.}
\vspace{-12mm}
\label{tab:ablation_contrib_module}
\end{table}

%\subsection{Qualitative Results}
%Figure ~\ref{fig:visualization_kitti_occ} shows the qualitative results achieved by our Fast Occupancy Network on the Semantic KITTI dataset. Our model can reconstruct the overview of the surrounding 3D scene well. The foreground objects can be well detected even if they are far away. However, the background at the edge of the image may be mushy, like vegetation and buildings, which may be related to the large area occlusion in the perspective view.

\section{Conclusion}
In this paper, we present a simple yet efficient Occupancy Network method, which leverages 2D deformable convolution to lift the BEV feature to the 3D voxel feature for occupancy prediction, achieving SOTA performance compared with other complex lifting methods with less computational cost. We present a PV segmentation supervision to improve performance without introducing any cost during the inference phase. Further, we present a Partial Voxel FPN module with few computational costs to improve accuracy significantly. Our method can be seamlessly applied to mainstream BEV models to transform them into Occupancy Network models.  Experiments show our method can surpass other SOTA methods on both accuracy and inference speed, validating the superiority of our method.

\bibliographystyle{splncs04}
\bibliography{main}
\end{document}